\documentclass[sigconf]{acmart}
\copyrightyear{2021} 
\acmYear{2021} 
\setcopyright{acmcopyright}\acmConference[CIKM '21]{Proceedings of the 30th ACM International Conference on Information and Knowledge Management}{November 1--5, 2021}{Virtual Event, QLD, Australia}
\acmBooktitle{Proceedings of the 30th ACM International Conference on Information and Knowledge Management (CIKM '21), November 1--5, 2021, Virtual Event, QLD, Australia}
\acmPrice{15.00}
\acmDOI{10.1145/3459637.3482000}
\acmISBN{978-1-4503-8446-9/21/11}

\usepackage{url}  
\usepackage{hyperref}
\usepackage{graphicx}  
\usepackage{balance}
\usepackage{booktabs}
\usepackage{multirow}
\usepackage{makecell}
\usepackage{subfigure}
\usepackage{mathtools}
\usepackage{enumerate}
\usepackage{bm}
\usepackage{amsmath} 

\begin{document}
\fancyhead{}
\title{DL-Traff: Survey and Benchmark of Deep Learning Models for Urban Traffic Prediction}

\author{Renhe Jiang$^{1, 2}$*, Du Yin$^{2}$*, Zhaonan Wang$^1$, Yizhuo Wang$^{2}$, Jiewen Deng$^2$,  Hangchen Liu$^{2}$, Zekun Cai$^{1}$, Jinliang Deng$^{2,3}$, Xuan Song$^{2, 1}\dagger$, Ryosuke Shibasaki$^{1}$}
\affiliation{$^1$The University of Tokyo
	\country{Japan}}
\affiliation{$^2$Southern University of Science and Technology
	\country{China}}
\affiliation{$^3$University of Technology Sydney
	\country{Australia}}
\thanks{* Equal contribution; $\dagger$ Corresponding author.}
\thanks{This work was supported by Grant-in-Aid for Early-Career Scientists (20K19859) of Japan Society for the Promotion of Science (JSPS).}
\email{jiangrh@csis.u-tokyo.ac.jp; yind7@outlook.com; songxuan@csis.u-tokyo.ac.jp}
\renewcommand{\shortauthors}{Jiang and Yin, et al.}


\begin{abstract}
Nowadays, with the rapid development of IoT (Internet of Things) and CPS (Cyber-Physical Systems) technologies, big spatiotemporal data are being generated from mobile phones, car navigation systems, and traffic sensors. By leveraging state-of-the-art deep learning technologies on such data, urban traffic prediction has drawn a lot of attention in AI and Intelligent Transportation System community. The problem can be uniformly modeled with a 3D tensor (T, N, C), where T denotes the total time steps, N denotes the size of the spatial domain (i.e., mesh-grids or graph-nodes), and C denotes the channels of information. According to the specific modeling strategy, the state-of-the-art deep learning models can be divided into three categories: grid-based, graph-based, and multivariate time-series models. In this study, we first synthetically review the deep traffic models as well as the widely used datasets, then build a standard benchmark to comprehensively evaluate their performances with the same settings and metrics. Our study named DL-Traff is implemented with two most popular deep learning frameworks, i.e., TensorFlow and PyTorch, which is already publicly available as two GitHub repositories \url{https://github.com/deepkashiwa20/DL-Traff-Grid} and \url{https://github.com/deepkashiwa20/DL-Traff-Graph}. With DL-Traff, we hope to deliver a useful resource to researchers who are interested in spatiotemporal data analysis.

\end{abstract}



\maketitle

\begin{table*}[t]
    \small
	\begin{tabular*}{17.7cm}{@{\extracolsep{\fill}}l|l|l|l|l|l}
		\hline
		\textbf{Grid-Based} & \textbf{Venue} & \textbf{Cite} & \textbf{Dataset (* means Open)} & \textbf{Prediction Task} & \textbf{Metric}\\
		\hline
		ST-ResNet\cite{zhang2017deep} & AAAI17 & 867 & TaxiBJ*, BikeNYC* & Taxi In-Out Flow & RMSE \\
		\hline
		DeepSD\cite{wang2017deepsd} & ICDE17 & 125 & Didi Taxi (HangZhou) & Taxi Demand & MAE, RMSE\\
		\hline
		DMVST-Net\cite{yao2018deep} & AAAI18 & 456 & Didi Taxi (GuangZhou) & Taxi Demand & RMSE, MAPE\\
		\hline
		Periodic-CRN\cite{zonoozi2018periodic} & IJCAI18 & 57 & TaxiBJ*, TaxiSG & Taxi Density/In-Out Flow & RMSE\\
		\hline
		Hetero-ConvLSTM\cite{yuan2018hetero} & KDD18 & 121 & Vehicle Crash Data* & Traffic Accident & MSE, RMSE, CE\\
		\hline
		DeepSTN+\cite{lin2019deepstn+} & AAAI19 & 53 & MobileBJ, BikeNYC-I* & Crowd/Taxi In-Out Flow & MAE, RMSE  \\
		\hline
		STDN\cite{yao2019revisiting} & AAAI19 & 204 & TaxiNYC*, BikeNYC-II* & Taxi/Bike O-D Number & RMSE, MAPE\\
		\hline
		MDL\cite{zhang2019flow} & TKDE19 & 92 & TaxiBJ, BikeNYC & Taxi Transition/In-Out Flow & MAE, RMSE\\
		\hline
		DeepUrbanEvent\cite{jiang2019deepurbanevent} & KDD19 & 32 & Konzatsu Toukei & Crowd Density/Flow & MSE\\
		\hline
		Curb-GAN \cite{zhang2020curb} & KDD20 & 1 & Taxi Speed/Inflow (Shenzhen)* & Taxi Speed/Inflow & RMSE, MAPE\\
		\toprule
		\hline
		\textbf{Graph-Based} & \textbf{Venue} & \textbf{Cite} & \textbf{Dataset (* means Open)} & \textbf{Prediction Task} & \textbf{Metric}\\
		\hline
		STGCN\cite{yu2018spatio} & IJCAI18 & 660 &  BJER4, PeMSD7(M)*, PeMSD7(L) & Traffic Volume/Speed & MAE, RMSE, MAPE\\
		\hline
		DCRNN(GCGRU)\cite{li2018diffusion} & ICLR18 & 691  & METR-LA*, PeMS-BAY* & Traffic Volume/Speed & MAE, RMSE, MAPE\\
		\hline
		Multi-graph\cite{chai2018bike} & SIGSPATIAL18 & 89 & Bike Flow (New York and Chicago) & Bike In-Out Flow & RMSE\\
		\hline
		ASTGCN\cite{guo2019attention}& AAAI19 & 239 & PeMSD4-I*, PeMSD8-I* & Traffic Volume/Speed & MAE, RMSE \\
		\hline
		DGCNN\cite{diao2019dynamic} & AAAI19 & 56 & TrafficNYC, PeMS & Traffic Volume/Speed & MAE, RMSE\\
		\hline
		ST-MGCN\cite{geng2019spatiotemporal} & AAAI19 & 182 & Bike Demand (Beijing and Shanghai) & Bike Demand & MAE, RMSE, MAPE\\
		\hline
		Graph WaveNet\cite{wu2019graph} & IJCAI19 & 144 & METR-LA*, PeMS-BAY* & Traffic Volume/Speed & MAE, RMSE, MAPE\\
		\hline
		STG2Seq\cite{bai2019stg2seq} &IJCAI19 & 33 & DidiSY, BikeNYC*, TaxiBJ* & Taxi/Bike Demand & MAE, RMSE, MAPE\\
		\hline
		T-GCN\cite{zhao2019t} &TITS19 & 195 & TaxiSZ*, METR-LA* & Traffic Volume/Speed & MAE, RMSE, Acc, $R^2$, var\\
		\hline
		TGC-LSTM\cite{cui2019traffic} & TITS19 & 166 & Seattle-Loop*, INRIX Traffic & Traffic Volume/Speed & MAE, RMSE, MAPE\\
		\hline
		GCGA\cite{yu2019real} & TITS19 & 27 & Cologne Traffic & Traffic Volume/Speed & MAPE\\
		\hline
		GMAN\cite{zheng2020gman} & AAAI20 & 73 & Taxi Xiamen, PeMS-BAY* & Traffic Volume/Speed & MAE, RMSE, MAPE\\
		\hline
		MRA-BGCN\cite{chen2020multi} & AAAI20 & 28 & METR-LA*, PeMS-BAY* & Traffic Volume/Speed & MAE, RMSE, MAPE\\
		\hline
		STSGCN\cite{song2020spatial} & AAAI20 & 39 & PeMS03*, PeMS04*, PeMS07*, PeMS08* & Traffic Volume/Speed & MAE, RMSE, MAPE\\
		\hline
		\multirow{2}{*}{SLCNN\cite{zhang2020spatio}} & \multirow{2}{*}{AAAI20} & \multirow{2}{*}{13} & METR-LA*, PeMS-BAY*, & \multirow{2}{*}{Traffic Volume/Speed} & \multirow{2}{*}{MAE, RMSE, MAPE} \\
		&  & & PeMSD7(M)*, BJF, BRF, BRF-L &   &  \\
		\hline
		STGNN\cite{wang2020traffic} & WWW20 & 24 & METR-LA*, PeMS-BAY* & Traffic Volume/Speed & MAE, RMSE, MAPE\\
		\hline
		H-STGCN\cite{dai2020hybrid} & KDD20 & 9 & W3-715, E5-2907 (Beijing) & Traffic Volume/Speed & MAE, MAPE, RMSE\\
		\hline
		AGCRN\cite{bai2020adaptive} & NeurIPS20 & 9 & PeMSD4*, PeMSD8* & Traffic Volume/Speed & MAE, RMSE, MAPE\\
		\hline
		T-MGCN\cite{lv2020temporal} & TITS20 & 7 & HZJTD*, PeMSD10*  &Traffic Volume/Speed & MAE, RMSE, MAPE\\
		\hline
		DGCN\cite{guo2020dynamic} & TITS20 & 1 & PeMSD4*, PeMSD8*, PHILADELPHIA &Traffic Volume/Speed & MAE, RMSE\\
		\toprule
		\hline
		\textbf{Multivariate Time-Series} & \textbf{Venue} & \textbf{Cite} & \textbf{Dataset (* means Open)} & \textbf{Prediction Task} & \textbf{Metric}\\
		\hline
		\multirow{2}{*}{LSTNet\cite{lai2018modeling}} & \multirow{2}{*}{SIGIR18} & \multirow{2}{*}{318} & PeMS-BAY*, Solar-Energy* & Multivariate Time-Series & \multirow{2}{*}{RSE, CORR}\\
		&&& Electricity*, Exchange Rate* & Traffic Volume/Speed & \\
		\hline
		\multirow{2}{*}{GaAN(GGRU)\cite{zhang2018gaan}} & \multirow{2}{*}{UAI18} & \multirow{2}{*}{214} & \multirow{2}{*}{PPI, Reddit, METR-LA*} & Node Classification & \multirow{2}{*}{MAE, RMSE}\\
		& & & & Traffic Volume/Speed & \\
		\hline
		GeoMAN\cite{liang2018geoman} & IJCAI18 & 182 & Water Quality, Air Quality & Multivariate Time-Series & MAE, RMSE \\
		\hline
		\multirow{2}{*}{ST-MetaNet\cite{pan2019urban}} & \multirow{2}{*}{KDD19} & \multirow{2}{*}{91} & \multirow{2}{*}{TaxiBJ-I*, METR-LA*} & Taxi In-Out Flow & \multirow{2}{*}{MAE, RMSE}\\
		& & & & Traffic Volume/Speed & \\
		\hline
		\multirow{2}{*}{TPA-LSTM \cite{shih2019temporal}} & \multirow{2}{*}{ECMLPKDD19} & \multirow{2}{*}{88} & Solar Energy*, Traffic(PeMS)*, , & Multivariate Time-Series & \multirow{2}{*}{RAE, RSE, CORR} \\
		& & & Electricity*, Music*, Exchange Rate* & Traffic Volume/Speed &\\
		\hline
		Transformer\cite{li2019enhancing} & NeurIPS19 & 84 & Electricity*, Traffic*, Solar*, Wind* & Multivariate Time-Series & $\rho$-quantile \\
		\hline
		\multirow{2}{*}{MTGNN\cite{wu2020connecting}} & \multirow{2}{*}{KDD20} & \multirow{2}{*}{32}& Solar-Energy*, Taffic(PeMS)*
		& Multivariate Time-Series & MAE, RMSE, MAPE\\
		& & & Electricity*, Exchange-Rate* & Traffic Volume/Speed & RSE, CORR\\
		\toprule
	\end{tabular*}
	\caption{Summary of The State-Of-The-Art Models *Citation number was referred from Google Scholar by 2021/6/13*}
	\label{tab:modelsummary}
\end{table*}

\section{Introduction}
Nowadays, with the rapid development of IoT (Internet of Things) and CPS (Cyber-Physical Systems) technologies, big spatiotemporal data are being generated from mobile phones, car navigation systems, and traffic sensors. Based on such data, urban traffic prediction has been taken as a significant research problem and a key technique for building smart city, especially intelligent transportation system. From 2014 to 2017, encouraged by the huge success of deep learning technologies in the Computer Vision and Natural Language Processing field, researchers in the Intelligent Transportation System community, started to apply Long-Term Short Memory (LSTM) and Convolution Neural Network (CNN) to the well-established traffic prediction task\cite{huang2014deep,lv2014traffic,ma2015long,ma2017learning}, and also achieved an unprecedented success. Following these pioneers, researchers have leveraged the state-of-the-art deep learning technologies to develop various prediction models and publish a big amount of studies on the major AI and transportation venues as listed in Table \ref{tab:modelsummary}. Although the prediction tasks may slightly differ from each other, they can all be categorized as deep traffic models.

\begin{figure}[h]
	\centering	
	\includegraphics[width=0.45\textwidth]{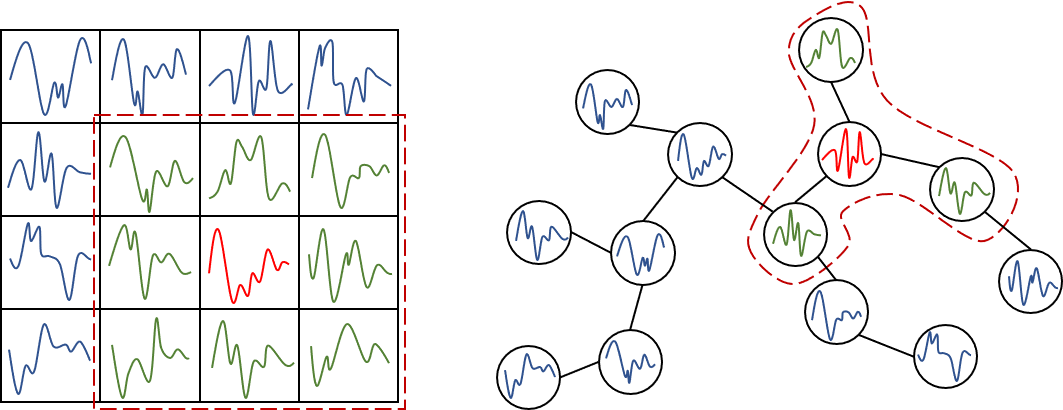}
	\caption{Grid-Based Traffic and Graph-Based Traffic.}
	\label{fig:intro}
\end{figure}

No matter based on grid or graph, the traffic data illustrated in Fig.\ref{fig:intro} can be uniformly represented with a 3D tensor $\mathbb{R}^{T\times N \times C}$, where T denotes the size of the temporal domain (i.e., timeslots with constant sampling rate), N denotes the size of the spatial domain (i.e., mesh-grids or graph-nodes), and C denotes the number of information channels. For instance, assuming 300 traffic sensors are deployed to record traffic speed (channel1) and volume (channel2) every 30 minutes for 100 consecutive days, then the total data can be represented by tensor $\mathbb{R}^{4800\times 300 \times 2}$. Besides traffic volume and speed, channels can also be used to store crowd density, taxi demand, traffic accident, car/ride-hailing order, and crowd/taxi/bike inflow and outflow. More specifically, grid-based model meshes the entire spatial domain into $H\times W$ fine-grained mesh-grids and converts the 3D representation into 4D tensor $\mathbb{R}^{T\times H\times W\times C}$ format. Graph-based model introduces directed or undirected graph $G$ = $(V,E)$ to utilize the topological structure of the urban road network for modeling, where $v\in V$ is a node, $|V|$ = $N$, and $e\in E$ is an edge. Multivariate time-series model naturally takes the N spatial units as N time-series variates and shares the same representation, i.e., $\mathbb{R}^{T\times N \times C}$ with graph-based model. Thus, the deep learning models listed in Table \ref{tab:modelsummary} can be divided into three groups according to the specific modeling strategy along the spatial axis.

Through the citation number in Table \ref{tab:modelsummary}, we can know how much attention these studies have drawn in our AI and data science community. But due to the huge amount of the related works, researchers are often too exhausted to follow up with the specific details of each model. More importantly, the evaluations on this family of models are still confusing and not well organized. For instance, some models demonstrated superior performances to the existing ones by using different datasets or metrics as shown in Table \ref{tab:modelsummary}, while some models utilized a self-designed objective function or employed extra data sources such as Point-of-Interest (POI) data \cite{lin2019deepstn+} or navigation app data \cite{dai2020hybrid} to achieve better prediction accuracy. To address the problems above, a concise but precise survey will be a great help for researchers involved in this emerging topic. But only a survey is not enough. It is also significant to conduct standard performance evaluations to examine the true function of each spatial and temporal component by using the same datasets, metrics, and other experimental settings. This paper fills these needs by providing a concise survey followed by a comprehensive benchmark evaluation on the recent deep traffic models. 

We first define two benchmark tasks in Section 2, one is single-step prediction for inflow and outflow based on grid-based traffic data, another is multi-step prediction for traffic speed based on graph-based data. Second, in Section 3, we investigate both of the grid-based and graph-based datasets and pick up some open and widely used ones as our benchmark data including TaxiBJ, BikeNYC, TaxiNYC, METR-LA, PeMS-BAY, and PeMSD7M. Next, in Section 4, we decompose the models into spatial and temporal units and give the roadmap that how the models evolve along the spatial and temporal axis. Further, we draw the architectures for a bunch of representative models (e.g., ST-ResNet\cite{zhang2017deep}, DMVST-Net\cite{yao2018deep}, STDN\cite{yao2019revisiting}, DeepSTN+\cite{lin2019deepstn+}, STGCN\cite{yu2018spatio}, DCRNN\cite{li2018diffusion}, Graph WaveNet\cite{wu2019graph}) in an intuitive and comparative manner. Then, in Section 5, we do a comprehensive evaluation on both the grid-based and graph-based models by using the benchmark tasks and datasets under the same settings and metrics (RMSE, MAE, MAPE). In Section 6, we briefly introduce the implementation details, the availability, and the usability of our benchmark. Finally, we give our conclusion in Section 7. 
The contributions of our work are summarized as follows:
\begin{itemize}
\item We give a concise but concrete survey on the recent deep traffic models. The technique detail and the evolution are clearly summarized along spatial and temporal axes.
\item We carefully select two traffic flow prediction tasks, four grid-based traffic datasets, and three graph-based traffic datasets, and implement plenty of grid/graph-based state-of-the-arts to form a complete benchmark called \textbf{DL-Traff}.
\item On this benchmark, we conduct a comprehensive evaluation of the effectiveness and efficiency performances of the-state-of-the-arts.
\item Our benchmark is implemented with the two most popular deep learning frameworks, i.e., TensorFlow and PyTorch. \textbf{DL-Traff} is already publicly available as two GitHub repositories \url{https://github.com/deepkashiwa20/DL-Traff-Grid} and \url{https://github.com/deepkashiwa20/DL-Traff-Graph}.
\end{itemize}
With DL-Traff, (1) users can quickly grasp the technical details about the state-of-the-art deep spatiotemporal models; (2) users can smoothly reproduce the prediction results reported in this paper and use them as the baselines; (3) users can easily launch a new deep solution with either TensorFlow or PyTorch for not only traffic flow prediction tasks, but also for other spatiotemporal problems such as anomaly/accident, electricity consumption, air quality, etc.

\section{Problem}
In this paper, we employ the following two prediction tasks into our benchmark. 
\begin{enumerate}
    \item Grid-based inflow and outflow prediction proposed by\cite{hoang2016forecasting,zhang2016dnn}. The problem is to predict how many taxis/bikes will flow into or out from each mesh-grid in the next time interval. It takes $\alpha$ steps of historical observations as input and gives the next step prediction as follows: [$X_{t-(\alpha-1)}$,...,$X_{t-1}$,$X_{t}$] $\rightarrow$ $X_{t+1}$, where $X_i$ $\in$ $\mathbb{R}^{H\times W\times C}$, $H,W$ are the indexes for the mesh, and C is equal to 2, respectively used for inflow and outflow.
    \item Graph-based traffic speed prediction as defined in \cite{yu2018spatio,li2018diffusion}. In order to make a variation to the first task, we define this task as multi-step-to-multi-step one as follows: [$X_{t-(\alpha-1)}$,...,$X_{t-1}$,$X_{t}$] $\rightarrow$ [$X_{t+1}$,$X_{t+2}$,...,$X_{t+\beta}$], where $X_i$ $\in$ $\mathbb{R}^{N\times C}$, $\alpha$/$\beta$ is the number of steps of observations/predictions, $N$ is the number of traffic sensors (i.e., nodes), and $C$ is equal to 1 that only stores the traffic speed.
\end{enumerate}
 
\begin{table*}[t]
    \small
	\centering
	\caption{Summary of The Public Traffic Datasets}
	\label{tab:datasummary}
	\begin{tabular*}{17.7cm}{@{\extracolsep{\fill}}l|l|l|l|l|l}
		\hline
		\textbf{Grid-Based} & \textbf{Reference} & \textbf{Data Description / Data Source} & \textbf{Spatial Domain} & \textbf{Time Period} & \textbf{Time Interval}\\
		\hline
		\multirow{2}{*}{TaxiBJ*}& \cite{zhang2017deep,zonoozi2018periodic} & \multirow{2}{*}{Taxi In-Out Flow / Taxi GPS Data of Beijing} & \multirow{2}{*}{32$\times$32 grids} & 2013/7/1$\sim$2016/4/10 & \multirow{2}{*}{30 minutes}\\
		& \cite{bai2019stg2seq} & & & *Four Time Periods & \\
		\hline
		TaxiBJ-I* &\cite{pan2019urban} & Taxi In-Out Flow / Taxi GPS Data of Beijing (TDrive) & 32$\times$32 grids & 2015/2/1$\sim$2015/6/2 & 60 minutes\\
		\hline
		\multirow{2}{*}{BikeNYC*} & \multirow{2}{*}{\cite{zhang2017deep,bai2019stg2seq}} & Bike In-Out Flow / Bike Trip Data of New York City  & \multirow{2}{*}{16$\times$8 grids} & \multirow{2}{*}{2014/4/1$\sim$2014/9/30} & \multirow{2}{*}{60 minutes}\\
		& & Citi Bike: \url{https://www.citibikenyc.com/system-data} && \\
		\hline
		BikeNYC-I* & \cite{lin2019deepstn+} & Bike In-Out Flow / Bike Trip Data of New York City & 21$\times$12 grids & 2014/4/1$\sim$2014/9/30 & 60 minutes \\
		\hline
		BikeNYC-II* & \cite{yao2019revisiting} & Bike In-Out Flow / Bike Trip Data of New York City & 10$\times$20 grids & 2016/7/1$\sim$2016/8/29 & 30 minutes\\
		\hline
		\multirow{3}{*}{TaxiNYC*} & \multirow{3}{*}{\cite{yao2019revisiting}} & Taxi In-Out Flow / Taxi Trip Data of New York City & \multirow{3}{*}{10$\times$20 grids} & \multirow{3}{*}{2015/1/1$\sim$2015/3/1} & \multirow{3}{*}{30 minutes}\\
		&& The New York City Taxi\&Limousine Commission && \\
		&& (TLC) \url{https://www1.nyc.gov/site/tlc/about/data.page} && \\
		\toprule
		\hline
		\textbf{Graph-Based} & \textbf{Reference} & \textbf{Data Description / Data Source} & \textbf{Spatial Domain} & \textbf{Time Period} & \textbf{Time Interval}\\
		\hline
		\multirow{4}{*}{METR-LA*}
		&\cite{wang2020traffic} &
		Traffic Speed Sensors in Los Angeles County & \multirow{4}{*}{207 sensors} & \multirow{4}{*}{2012/3/1$\sim$2012/6/30} & \multirow{4}{*}{5 minutes}\\
		&\cite{zhao2019t,li2018diffusion}& Los Angeles Metropolitan Transportation Authority& & &\\
		&\cite{wu2019graph,zhang2018gaan}& *Collaborated with University of Southern California& & &\\
		&\cite{chen2020multi,pan2019urban}& \url{https://imsc.usc.edu/platforms/transdec/} & & &\\
		\hline
		\multirow{3}{*}{PeMS-BAY*} & \cite{li2018diffusion,lai2018modeling} & Traffic Speed Sensors in California & \multirow{3}{*}{325 sensors} & \multirow{3}{*}{2017/1/1$\sim$2017/5/31} & \multirow{3}{*}{5 minutes}\\
		&\cite{wu2019graph,chen2020multi}& Caltrans Performance Measurement System (PeMS) & &\\
		&\cite{wang2020traffic,zheng2020gman}& PeMS: \url{http://pems.dot.ca.gov/} & & &\\
		\hline
		PeMSD7(M)* & \cite{yu2018spatio} & Traffic Speed Sensors in California (PeMS) & 228 sensors& 2012/5/1$\sim$2012/6/30 & 5 minutes\\
		PeMS03* &\cite{song2020spatial} & Traffic Speed Sensors in California (PeMS) & 358 sensors & 2018/9/1$\sim$2018/11/30& 5 minutes\\
		PeMSD4(PeMS04)* &\cite{song2020spatial,bai2020adaptive} & Traffic Speed Sensors in California (PeMS) & 307 sensors & 2018/1/1$\sim$2018/2/28& 5 minutes\\
		PeMS07* &\cite{song2020spatial} & Traffic Speed Sensors in California (PeMS) & 883 sensors & 2017/5/1$\sim$2017/8/31& 5 minutes\\
		PeMSD8(PeMS08)* &\cite{song2020spatial,bai2020adaptive} & Traffic Speed Sensors in California (PeMS) & 170 sensors & 2016/7/1$\sim$2016/8/31& 5 minutes\\
		PeMSD4-I* &\cite{guo2019attention} & Traffic Speed Sensors in California (PeMS) & 3848 sensors & 2018/1/1$\sim$2018/2/28 & 5 minutes\\
		PeMSD8-I* &\cite{guo2019attention} & Traffic Speed Sensors in California (PeMS) & 1979 sensors & 2016/7/1$\sim$2016/8/31& 5  minutes\\
		PeMSD10* &\cite{lv2020temporal} & Traffic Speed Sensors in California (PeMS) &  608 sensors &2018/1/1$\sim$2018/3/31& 15 minutes\\
		Traffic(PeMS)* &\cite{shih2019temporal,wu2020connecting} & Traffic Speed Sensors in California (PeMS) & 862 sensors & 2015/1/1$\sim$2016/12/31& 60 minutes\\
		\hline
		LOOP-SEATTLE* &\cite{cui2019traffic} & Traffic Speed Sensors in Greater Seattle Area & 323 sensors & 2015/1/1$\sim$2015/12/31& 5 minutes\\
		\hline
		TaxiSZ* &\cite{zhao2019t} & Taxi Speed on Roads / Taxi GPS Data of Shenzhen & 156 roads & 2015/1/1$\sim$2015/1/31 & 15 minutes \\
		\hline
		\multirow{2}{*}{HZJTD*} & \multirow{2}{*}{\cite{lv2020temporal}} & Traffic Speed Sensors in Hangzhou & \multirow{2}{*}{202 sensors} & \multirow{2}{*}{2013/10/16$\sim$2014/10/3} & \multirow{2}{*}{15 minutes}\\
		& & Hangzhou Integrated Transportation Research Center & &\\
		\toprule
		
		
	\end{tabular*}
\end{table*}


\section{Dataset}
The public datasets for urban traffic prediction are summarized in Table \ref{tab:datasummary}, where the reference, source, and spatial and temporal spec are enumerated. We pick up some widely used ones as our benchmark datasets.
\begin{figure}[h]
	\centering
	\includegraphics[width=0.49\textwidth]{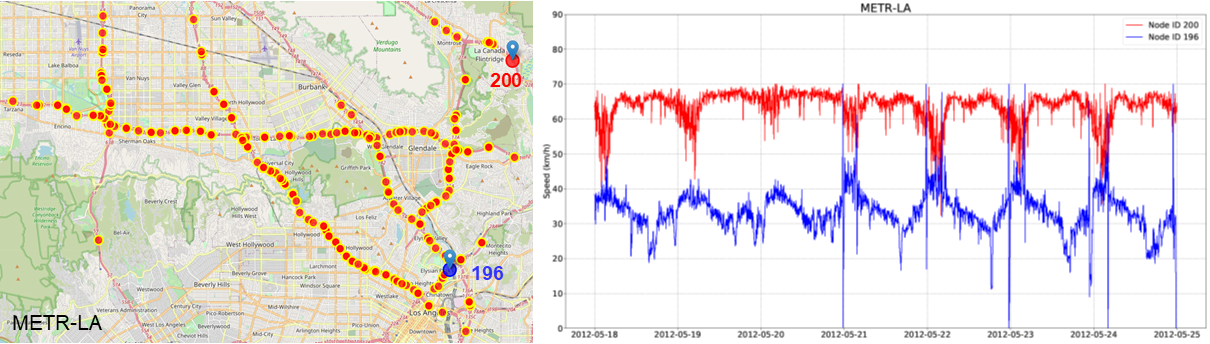}
	\caption{Visualization of METR-LA.}
	\label{fig:dataset_one}
\end{figure}
\subsection{Grid-Based Traffic Dataset}
\noindent\textbf{TaxiBJ.} This is taxi in-out flow data published by \cite{zhang2017deep}, created from the taxicab GPS data in Beijing from four separate time periods: 2013/7/1-2013/10/30, 2014/3/1-2014/6/30, 2015/3/1-2015/6/30, and 2015/11/1-2016/4/10. Based on the same underlying taxi GPS data (T-Drive), a similar dataset denoted as TaxiBJ-I is created by \cite{pan2019urban}.

\noindent\textbf{BikeNYC.} This is bike in-out flow data of New York City from 2014/4/1 to 2014/9/30 used by \cite{zhang2017deep}. The original bike trip data is published by Citi Bike, NYC's official bike-sharing system, which includes: trip duration, starting and ending station IDs, and start and end times. Similar datasets \textbf{BikeNYC-I}, \textbf{BikeNYC-II} were used by \cite{lin2019deepstn+} and \cite{yao2019revisiting} respectively. These two will be used in our experiment due to the larger spatial domain.

\noindent\textbf{TaxiNYC.} This is taxi in-out flow data of New York City from 2015/1/1~2015/3/1 used by \cite{yao2019revisiting}. The original taxi trip data is published by the New York City Taxi and Limousine Commission (TLC), that includes pick-up and drop-off dates/times, pick-up and drop-off locations, trip distances, itemized fares, driver-reported passenger counts, etc.

\subsection{Graph-Based Traffic Dataset}
\noindent\textbf{METR-LA.} This is a Los Angeles traffic data published by \cite{li2018diffusion}. The data are collected from 207 highway sensors within 4 months from 2012/3/1 to 2012/6/30. A quite number of studies used this dataset as shown in Table \ref{tab:datasummary}. To be intuitive, a data visualization has been made as Fig.\ref{fig:dataset_one}.

\noindent\textbf{PeMS-BAY.} This is a traffic flow dataset collected from California Transportation Agencies Performance Measurement System (PeMS). It contains 325 traffic sensors in the Bay Area from 2017/1/1 to 2017/5/31. Massive studies also generate a variety of PeMS datasets by using the same source. 

\noindent\textbf{PeMSD7M.} This traffic dataset is created and published by \cite{yu2018spatio}, also collected from PeMS. It covers 228 traffic sensors lasting from 2012/5/1 to 2012/6/30 with a 5-minute sampling rate on weekdays.

\noindent\textbf{Summary.} The taxi and bike trip data published by Citi Bike and TLC of New York City and the traffic sensor data from PeMS of California are taken as three trustworthy and wildly-used data sources for traffic prediction. Researchers can easily access the data through the URLs listed in Table \ref{tab:datasummary}.



\begin{table*}[t]
    \small
	\centering
	\caption{Base Technologies Employed for Spatial and Temporal Modeling}
	\label{tab:basetechnology}
	\begin{tabular*}{17.8cm}{@{\extracolsep{\fill}}l|lll|llll|l|lll|llll}
		\hline
		& \multicolumn{3}{c|}{Spatial Axis} & \multicolumn{4}{c|}{Temporal Axis} & & \multicolumn{3}{c|}{Spatial Axis} & \multicolumn{4}{c}{Temporal Axis}\\
		\hline
		models & CNN & GCN & Attn. & LSTM & GRU & TCN & Attn. & models & CNN & GCN & Attn. & LSTM & GRU & TCN & Attn. \\
		\hline
		ST-ResNet\cite{zhang2017deep} & \checkmark &&&&&&& STGCN\cite{yu2018spatio} & & \checkmark & & & & \checkmark & \\ 
		DMVST-Net\cite{yao2018deep} & \checkmark &&& \checkmark &&&& GaAN(GGRU)\cite{zhang2018gaan} & & \checkmark&\checkmark &&\checkmark &&\\
		STDN\cite{yao2019revisiting} & \checkmark &&& \checkmark &&& \checkmark & DCRNN(GCGRU)\cite{li2018diffusion} & &\checkmark &&& \checkmark && \\
		DeepSTN+\cite{lin2019deepstn+} & \checkmark &&&&&& & Multi-graph\cite{chai2018bike} &&\checkmark&&\checkmark&&&\\
		LSTNet\cite{lai2018modeling} & \checkmark & & & & \checkmark & \checkmark& \checkmark & ASTGCN\cite{guo2019attention} && \checkmark &\checkmark &&&&\checkmark \\
		GeoMAN\cite{liang2018geoman} & & &\checkmark &\checkmark & & &\checkmark & TGCN\cite{zhao2019t} && \checkmark &&&\checkmark&&\\
		TPA-LSTM\cite{shih2019temporal} &&& & \checkmark & & \checkmark &\checkmark & Graph WaveNet\cite{wu2019graph} && \checkmark &&&& \checkmark &\\
		Transformer\cite{li2019enhancing} &&& & & & &\checkmark & MTGNN\cite{wu2020connecting} && \checkmark &&&& \checkmark &\\
		ST-MetaNet\cite{pan2019urban} &&&\checkmark &&\checkmark&&& STGNN\cite{wang2020traffic} &&\checkmark &\checkmark &&\checkmark&&\checkmark\\
		GMAN\cite{zheng2020gman} &&&\checkmark &&&&\checkmark &AGCRN\cite{bai2020adaptive} & &\checkmark &&& \checkmark & &\\
		\hline
	\end{tabular*}
\end{table*}

\section{Model}
Complex spatial and temporal dependencies are the key challenges in urban traffic prediction tasks.
Temporally, future prediction depends on the recent observations as well as the past periodical patterns; Spatially, the traffic states in certain mesh-grid or graph-node are affected by the nearby ones as well as distant ones. To capture the temporal dependency, LSTM\cite{hochreiter1997long} and its simplified variant GRU\cite{chung2014empirical} are respectively utilized by the models as shown in Table \ref{tab:basetechnology}. In parallel with the RNNs, 1D CNN and its enhanced version TCN \cite{yu2015multi} are also employed as the core technology for temporal modeling, and demonstrate the superior time efficiency and matchable effectiveness to LSTM and GRU. 

On the other hand, to capture the spatial dependency, grid-based models simply use the normal convolution operation\cite{lecun1998gradient} thanks to the natural euclidean property of grid spacing; graph-based models leverage the graph convolution in non-euclidean space \cite{defferrard2016convolutional,kipf2016semi} by involving the adjacency relation $A\in$ $\mathbb{R}^{N*N}$ between each pair of spatial units. Meanwhile, attention mechanism\cite{vaswani2017attention} also known as Transformer has rapidly taken over the AI community from natural language (GPT-3) to vision since 2020. Thus, attention is also introduced as base technology for modeling both spatial and temporal dependencies. 

\begin{figure*}[h]
	\centering
	\includegraphics[width=0.97\textwidth]{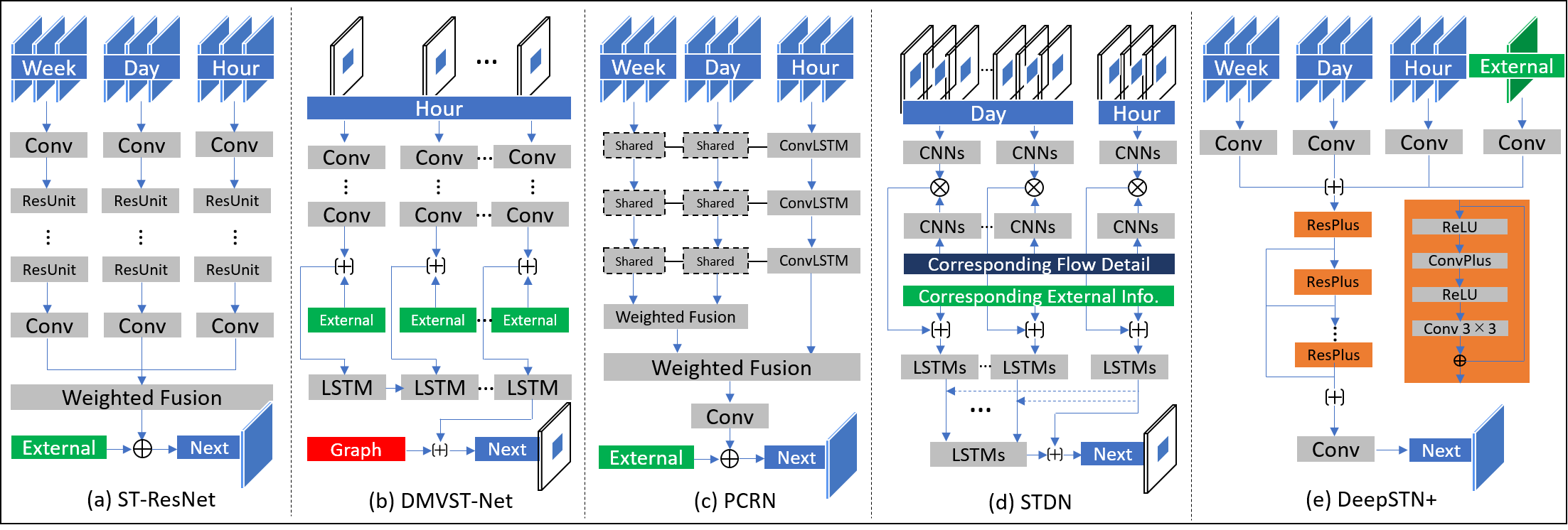}
	\caption{Architectures of Representative Grid-Based Models.}
	\label{fig:architectures}
\end{figure*}

\begin{figure*}[h]
	\centering
	\includegraphics[width=0.97\textwidth]{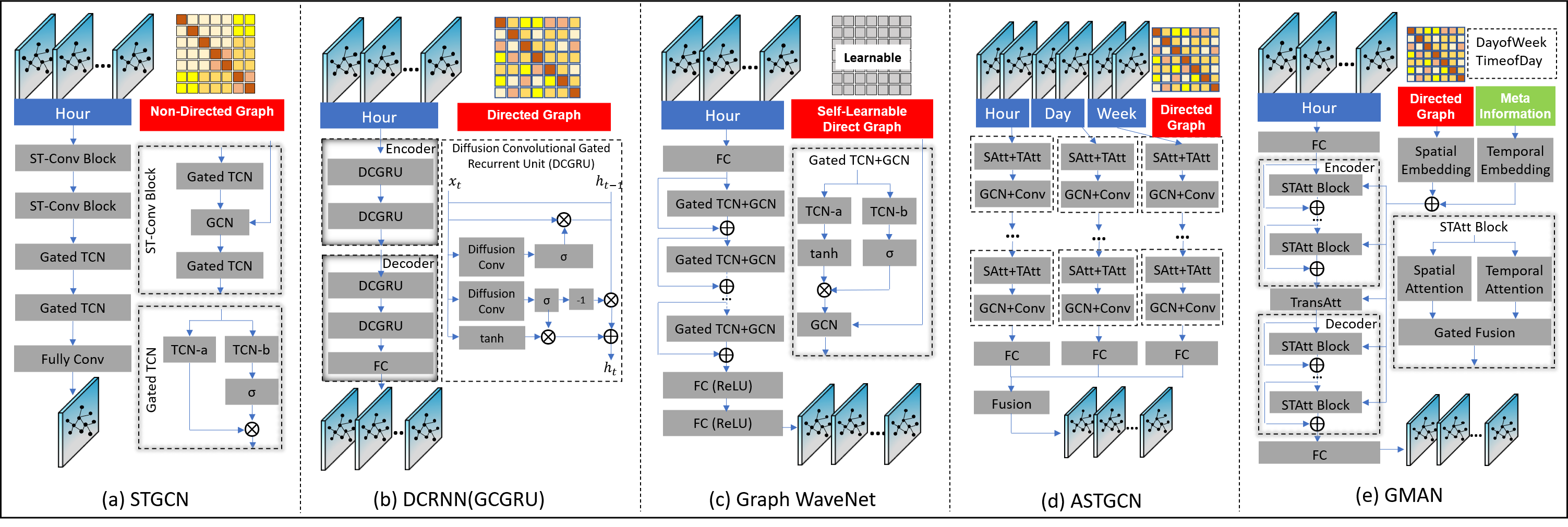}
	\caption{Architectures of Representative Graph-Based Models.}
	\label{fig:architectures1}
\end{figure*}

We select the most representative models in Table \ref{tab:modelsummary} and summarize the base technologies employed by each model for spatial and temporal modeling as Table \ref{tab:basetechnology}. On the other hand, for better understanding, we simplify and plot the network architectures in a unified manner for five grid-based models including
ST-ResNet\cite{zhang2017deep}, DMVST-Net\cite{yao2018deep}, Periodic-CRN(PCRN)\cite{zonoozi2018periodic}, STDN\cite{yao2019revisiting}, and DeepSTN+\cite{lin2019deepstn+} as Fig.\ref{fig:architectures}, and five graph-based models, namely STGCN\cite{yu2018spatio}, DCRNN\cite{li2018diffusion}, Graph WaveNet\cite{wu2019graph}, ASTGCN\cite{guo2019attention}, and GMAN\cite{zheng2020gman} as Fig.\ref{fig:architectures1}. Through Fig.\ref{fig:architectures}$\sim$\ref{fig:architectures1}, we can easily understand how the spatial and temporal modules listed in Table \ref{tab:basetechnology} are assembled to form an integrated model. 

Moreover, we describe how the employed technologies are evolving along the spatial and temporal axis for both grid-based and graph-based models in the next two subsections. Note that the multivariate time-series (MTS) models such as LSTNet\cite{lai2018modeling}, TPA-LSTM\cite{shih2019temporal}, GeoMAN\cite{liang2018geoman}, and Transformer\cite{li2019enhancing} are also gradually evolving along the spatial and temporal axis. From the spatial perspective, they focus on correlation/dependence between variates; from the temporal perspective, they aim to utilize the periodic patterns occurred in time series. But due to space limitations, we don't expand the details of those MTS models in this paper.


\subsection{Roadmap for Grid-Based Model}
ST-ResNet\cite{zhang2017deep} is the earliest and the most representative grid-based deep learning method for traffic in-out flow prediction. It converts 4D tensor ($T$,$H$,$W$,$C$) into 3D tensor ($H$,$W$,$T$*$C$) by concatenating the channels at each time step so that CNN can be used to capture spatial dependency similarly to an image. Then, it creatively proposes a set of temporal features called $Closeness$, $Period$, and $Trend$, which correspond to \emph{the most recent observations}, \emph{daily periodicity}, and \emph{weekly trend} respectively. Intuitively, the three parts of the features can be represented by:

$X^{Closeness}$ = [$X_{t-l_c}$, $X_{t-(l_c-1)}$, ..., $X_{t-1}$]

$X^{Period}$ = [$X_{t-l_p \times s_p}$, $X_{t-(l_p-1) \times s_p}$, ..., $X_{t-s_p}$]

$X^{Trend}$ = [$X_{t-l_q \times s_q}$, $X_{t-(l_q-1) \times s_q}$, ..., $X_{t-s_q}$]

\noindent where $l_c$, $l_p$, $l_q$ are the sequence length of \{$Closeness$, $Period$, $Trend$\}, $s_p$ and $s_q$ are the time span of $Period$ and $Trend$, the $Closeness$ span $s_c$ is equal to 1 by default. This feature is not only inherited by the later grid-based models including STDN\cite{yao2019revisiting} and DeepSTN+\cite{lin2019deepstn+}, but also some graph-based models like ASTGCN\cite{guo2019attention}, which is still 
regarded as the state-of-the-art temporal feature by now. To capture the long-range spatial dependency between mesh-grids, it employs Residual Learning to construct deep enough CNN networks.  Additionally, it further utilizes external information including weather, event, and metadata(i.e. DayOfWeek, WeekdayOrWeekend) to auxiliarily enhance spatiotemporal modeling. 

\noindent\textbf{Improvement along Spatial Axis.} Different from ST-ResNet that takes the entire mesh-grids as input, DMVST-Net\cite{yao2018deep} and STDN\cite{yao2019revisiting} take one grid and its surrounding grids (i.e. $S$$\times$$S$ region) as input, thus a local CNN is enough to capture spatial dependency only among nearby grids. For the global spatial dependency, DMVST-Net introduces a weighted graph as an extra input, where nodes are the grids, and each edge represents the similarity of two time-series values (i.e. historical taxi demand) between any two grids. The graph will be manually embedded into a feature vector so that it can be concatenated with the other part. Through this, DMVST-Net gains the ability to capture long-range spatial dependency. Furthermore, STDN and \cite{zhang2019flow,jiang2019deepurbanevent} consider the local flow information (i.e. flow from one central grid to its surrounding $S$$\times$$S$ grids) to facilitate predicting the traffic volume in the central grid, which is implemented with a flow gating mechanism in STDN and multitask learning in \cite{zhang2019flow,jiang2019deepurbanevent}. DeepSTN+ \cite{lin2019deepstn+} uses Point-Of-Interest (POI) data as external information (e.g., office/residential/shopping area) to take the influence of location function on the crowd/traffic flow into consideration.

\noindent\textbf{Improvement along Temporal Axis.}
One major drawback of ST-ResNet is it does not explicitly handle the temporal axis, because it forces the video-like tensor ($T$,$H$,$W$,$C$) to be converted into an image-like tensor ($H$,$W$,$T$*$C$). To address this, DMVST-Net and STDN employ LSTM to connect with a separate and unshared CNN for each timestamp. STDN further considers the temporal shifting problem about periodicity (i.e. traffic data is not strictly periodic) and designs a \emph{Periodically Shifted Attention Mechanism} to solve the issue. Specifically, it sets a small time window to collect $Q$ time intervals right before and after the currently-predicting one. And the attention is used to obtain a weighted average representation $h$ from the $Q$ representations \{$h_1$, $h_2$, $...$, $h_Q$\} generated by LSTM. To this end, LSTM, and CNN work together to separately and sequentially model the spatial and temporal dependency. Convolutional LSTM \cite{xingjian2015convolutional} extends the fully connected LSTM (FC-LSTM) to have convolutional structures in both the input-to-state and state-to-state transitions and achieves a lot of successes on video modeling tasks. Motivated by this, ConvLSTM and its variant ConvGRU are utilized by \cite{zonoozi2018periodic,yuan2018hetero,jiang2019deepurbanevent} to simultaneously capture the spatial and temporal dependency.

\begin{table*}[h]
    \small
	\centering
	\caption{Performance Evaluation for Single-Step Prediction on Grid-Based Traffic Datasets}
	\label{tab:performance_grid}
	\begin{tabular*}{17.0cm}{@{\extracolsep{\fill}}l|ccc|ccc|ccc|ccc}
		\hline
		\multicolumn{1}{l|}{} &
		\multicolumn{3}{c|}{TaxiBJ} &
		\multicolumn{3}{c|}{BikeNYC-I} &
		\multicolumn{3}{c|}{BikeNYC-II} &
		\multicolumn{3}{c}{TaxiNYC} 
		\\
		\hline
		\multicolumn{1}{l|}{Model} & 
		\multicolumn{1}{c}{RMSE} & 
		\multicolumn{1}{c}{MAE} &
		\multicolumn{1}{c|}{MAPE} &
		\multicolumn{1}{c}{RMSE} & 
		\multicolumn{1}{c}{MAE} &
		\multicolumn{1}{c|}{MAPE} &
		\multicolumn{1}{c}{RMSE} & 
		\multicolumn{1}{c}{MAE} &
		\multicolumn{1}{c|}{MAPE} &
		\multicolumn{1}{c}{RMSE} & 
		\multicolumn{1}{c}{MAE} &
		\multicolumn{1}{c}{MAPE}
		\\
		\hline
		HistoricalAverage & 45.004 & 24.475 & 8.04\% & 15.676 & 4.882 & 5.45\% & 4.874 & 1.500 & 3.30\% & 21.535 & 7.121 & 4.56\%\\
		CopyLastStep & 23.609 & 13.372 & 6.20\% & 14.152 & 4.344 & 5.01\% & 4.999 & 1.606 & 3.50\% & 18.660 & 6.497 & 4.91\%\\
		CNN\cite{lecun1998gradient} & 23.550 & 13.797 & 8.46\% & 12.064 & 4.088 & 5.82\% & 4.511 & 1.574 & 3.98\% & 16.741 & 6.884 & 8.08\%\\
		ConvLSTM\cite{xingjian2015convolutional} & 19.247 & 10.816 & 5.61\% & 6.616 & 2.412 & 3.90\% & 3.174 & 1.133 & 2.90\% & 12.143 & 4.811 & 5.16\%\\
		ST-ResNet\cite{zhang2017deep} & 18.702 & 10.493 & 5.19\% & 6.106 & 2.360 & 3.72\% & 3.191 & 1.169 & 2.86\% & 11.553 & 4.535 & 4.32\%\\
		DMVST-Net\cite{yao2018deep} & 20.389 & 11.832 & 5.99\% & 7.990 & 2.833 & 3.93\% & 3.521 & 1.287 & 2.97\% & 13.605 & 4.928 & 4.49\%\\
		PCRN\cite{zonoozi2018periodic} & 18.629 & 10.432 & 5.45\% & 6.680 & \textbf{2.351} & 3.63\% & 3.149 & \textbf{1.107} & 2.78\% & 12.027 & 4.606 & 4.62\%\\
		DeepSTN+\cite{lin2019deepstn+} & 18.141 & 10.126 & 5.14\% & 6.205 & 2.489 & 3.48\% & 3.205 & 1.245 & 2.80\% & 11.420 & \textbf{4.441} & 4.45\%\\
		STDN\cite{yao2019revisiting} & \textbf{17.826} & \textbf{9.901} & \textbf{4.81\%} & \textbf{5.783} & 2.410 & \textbf{3.35\%} & \textbf{3.004} & 1.167 & \textbf{2.67\%} & \textbf{11.252} & 4.474 & \textbf{4.09\%}\\
		\hline
	\end{tabular*}
\end{table*}
\begin{table*}[h]
    \small
	\centering
	\caption{Performance Evaluation for Multi-Step Prediction on Graph-Based Traffic Datasets}
	\label{tab:trafficvolumn}
	\begin{tabular*}{17.0cm}{@{\extracolsep{\fill}}l|l|ccc|ccc|ccc}
		\hline
		\multicolumn{1}{l|}{} & \multicolumn{1}{l|}{} & \multicolumn{3}{c|}{3 Steps / 15 Minutes Ahead} &
		\multicolumn{3}{c|}{6 Steps / 30 Minutes Ahead} &
		\multicolumn{3}{c}{12 Steps / 60 Minutes Ahead}
		\\
		\hline
		\multicolumn{1}{l|}{Dataset} & \multicolumn{1}{l|}{Model} & 
		\multicolumn{1}{c}{RMSE} & 
		\multicolumn{1}{c}{MAE} &
		\multicolumn{1}{c|}{MAPE} & 
		\multicolumn{1}{c}{RMSE} & 
		\multicolumn{1}{c}{MAE} & 
		\multicolumn{1}{c|}{MAPE} & 
		\multicolumn{1}{c}{RMSE} & 
		\multicolumn{1}{c}{MAE} & 
		\multicolumn{1}{c}{MAPE} 
		\\
		\hline
	    \multirow{8}{*}{METR-LA} &  
		HistoricalAverage & 14.737 & 11.013 & 23.34\% & 14.737 & 11.010 & 23.34\% & 14.736 & 11.005 & 23.33\% \\
		
		& CopyLastSteps & 14.215 & 6.799 & 16.73\% & 14.214 & 6.799 & 16.73\% & 14.214 & 6.798 & 16.72\% \\

		& LSTNet\cite{lai2018modeling} & 8.067 & 3.914 & 9.27\% & 10.181 & 5.219 & 12.22\% & 11.890 & 6.335 & 15.38\% \\
		
		& STGCN\cite{yu2018spatio} & 7.918 & 3.469 & 8.57\% & 9.948 & 4.263 & 10.70\% & 11.813 & 5.079 & 13.09\% \\
		
		& DCRNN\cite{li2018diffusion} & \textbf{7.509} & 3.261 & 8.00\% & 9.543 & 4.021 & 10.12\% & 11.854 & 5.080 & 13.08\% \\
		
		& Graph WaveNet\cite{wu2019graph} & 7.512 & \textbf{3.204} & \textbf{7.62\%} &\textbf{9.445}  & \textbf{3.922} & \textbf{9.52\%} & \textbf{11.485} & \textbf{4.848} & \textbf{11.93\%} \\
		
		& ASTGCN\cite{guo2019attention} & 7.977 & 3.624 & 9.13\% & 10.042 & 4.514 & 11.57\% & 12.092 & 5.776 & 14.85\% \\
		
		& GMAN\cite{zheng2020gman} & 8.869 & 4.139 & 10.88\% & 9.917 & 4.517 & 11.77\% & 11.910 & 5.475 & 14.10\% \\

		& MTGNN\cite{wu2020connecting} & 7.707 & 3.277 & 8.02\% & 9.625 & 3.999 & 10.00\% & 11.624 & 4.867 & 12.17\% \\

		& AGCRN\cite{bai2020adaptive} & 7.558 & 3.292 & 8.17\% & 9.499 & 4.016 & 10.16\% & 11.502 & 4.901 & 12.43\% \\
		
		\hline
		\multirow{8}{*}{PeMS-BAY} &
		HistoricalAverage & 6.687 & 3.333 & 8.10\% & 6.686 & 3.333 & 8.10\% & 6.685 & 3.332 & 8.10\% \\
		
		& CopyLastSteps & 7.022 & 3.052 & 6.84\% & 7.016 & 3.049 & 6.84\% & 7.05 & 3.044 & 6.83\% \\

		& LSTNet\cite{lai2018modeling} & 3.224 & 1.643 & 3.47\% & 4.375 & 2.383 & 5.04\% & 5.515 & 2.974 & 6.86\% \\
		
		& STGCN\cite{yu2018spatio} & 2.827 & 1.327 & 2.79\% & 3.887 & 1.698 & 3.81\% & 4.748 & 2.055 & 5.02\% \\
		
		& DCRNN\cite{li2018diffusion} & 2.867 & 1.377 & 2.96\% & 3.905 & 1.726 & 3.97\% & 4.798 & 2.091 & 4.99\% \\
		
		& Graph WaveNet\cite{wu2019graph} & \textbf{2.759} & \textbf{1.322} & \textbf{2.78\%} & \textbf{3.737} & 1.660 & \textbf{3.75\%} & 4.562 & 1.991 & 4.75\% \\
		
		& ASTGCN\cite{guo2019attention} & 3.057 & 1.435 & 3.25\% & 4.066 & 1.795 & 4.40\% & 4.770 & 2.103 & 5.30\% \\
		
		& GMAN\cite{zheng2020gman} & 4.219 & 1.802 & 4.47\% & 4.143 & 1.794 & 4.40\% & 5.034 & 2.186 & 5.29\% \\

		& MTGNN\cite{wu2020connecting} & 2.849 & 1.334 & 2.84\% & 3.800 & \textbf{1.658} & 3.77\% & \textbf{4.491} & \textbf{1.950} & \textbf{4.59\%} \\

		& AGCRN\cite{bai2020adaptive} & 2.856 & 1.354 & 2.94\% & 3.818 & 1.670 & 3.84\% & 4.570 & 1.964 & 4.69\% \\
		
		\hline
		\multirow{8}{*}{PEMSD7M} &   
		HistoricalAverage & 7.077 & 3.917 & 9.90\% & 7.083 & 3.920 & 9.92\% & 7.095 & 3.925 & 9.95\% \\
		
		& CopyLastSteps & 9.591 & 5.021 & 12.33\% & 9.594 & 5.022 & 12.33\% & 9.597 & 5.024 & 12.34\% \\

		& LSTNet\cite{lai2018modeling} & 4.308 & 2.423 & 5.73\% & 8.951 & 5.132 & 12.22\% & 10.881 & 6.624 & 16.72\% \\
		
		& STGCN\cite{yu2018spatio}  & 4.051 & 2.124 & 5.02\% & 5.532 & 2.783 & 6.96\% & 6.695 & 3.374 & 8.74\% \\
		
		& DCRNN\cite{li2018diffusion} & 4.143 & 2.213 & 5.33\% & 5.679 & 2.907 & 7.41\% & 7.138 & 3.670 & 9.81\%  \\
		
		& Graph WaveNet\cite{wu2019graph} & \textbf{3.992} & 2.130 & \textbf{5.00\%} & \textbf{5.332} & 2.715 & 6.75\% & \textbf{6.431} & 3.266 & 8.47\% \\
		
		& ASTGCN\cite{guo2019attention} & 4.257 & 2.340 & 5.83\% & 5.506 & 2.992 & 7.69\% & 6.587 & 3.572 & 9.48\% \\
		
		& GMAN\cite{zheng2020gman} & 5.711 & 2.877 & 7.25\% & 6.171 & 3.084 & 7.77\% & 7.897 & 3.988 & 10.02\% \\

		& MTGNN\cite{wu2020connecting} & 4.032 & \textbf{2.120} & 5.02\% & 5.373 & \textbf{2.687} & \textbf{6.70\%} & 6.496 & \textbf{3.204} & \textbf{8.24\%} \\

		& AGCRN\cite{bai2020adaptive} & 4.073 & 2.167 & 5.19\% & 5.479 & 2.769 & 6.89\% & 6.733 & 3.358 & 8.55\% \\
		
		\hline
	\end{tabular*}
\end{table*}

\subsection{Roadmap for Graph-Based Model}
STGCN\cite{yu2018spatio} is one of the earliest models that use graph neural networks to predict traffic flow. Temporally, instead of RNN, it uses TCN \cite{yu2015multi} with a gated mechanism as shown in Fig.\ref{fig:architectures1} to capture the dependency only from $Closeness$ features. Spatially, it applies two spectral graph convolution, ChebyNet\cite{defferrard2016convolutional} and 1st-order approximation of ChebyNet \cite{kipf2016semi}. TCN and GCN are stacked together as an ST-Conv block to sequentially do the spatial and temporal modeling. One major limitation of STGCN is that it uses a symmetrical adjacency matrix (i.e., undirected graph) that considers the euclidean distance between two road sensors. Thus it is difficult to model the difference of the two-way traffic flow in one road. DCRNN \cite{li2018diffusion} is another pioneer to utilize graph convolution for traffic flow prediction. In contrast to the spectral convolution in STGCN, DCRNN applies a spatial graph convolution called Diffusion Convolution implemented with bidirectional random walks on a directed graph (i.e., non-symmetric adjacent matrix), so that it can capture the spatial influence from both the upstream and the downstream traffic flows. For the temporal axis, similar to ConvLSTM, it replaces the normal matrix multiplication in GRU with the proposed diffusion convolution, then a Diffusion Convolution Gated Recurrent Unit (DCGRU) is assembled that can simultaneously do the spatial and temporal modeling. With this DCGRU, it further implements an encoder-decoder structure to enable the multi-step-to-multi-step prediction. Inspired by STGCN and DCRNN, massive graph-based traffic models have been proposed as summarized in Table \ref{tab:modelsummary}.

\noindent\textbf{Improvement along Temporal Axis.} For the temporal feature, ASTGCN\cite{guo2019attention} inherits $Closeness$, $Period$, and $Trend$ from ST-ResNet, and improves STGCN that only takes $Closeness$. Besides, STSGCN\cite{song2020spatial} constructs a localized temporal graph by connecting all nodes with themselves at the previous and the next steps, updating the adjacency matrix from $A$$\in$$\mathbb{R}^{N*N}$ to $A'$$\in$$\mathbb{R}^{3N*3N}$, then only uses GCN to simultaneously do the spatial and temporal modeling. On the other hand, to get better ability of temporal modeling, T-GCN\cite{zhao2019t} and TGC-LSTM\cite{cui2019traffic} respectively use GRU and LSTM instead of TCN to improve STGCN; GCGA\cite{yu2019real} combines Generative Adversarial Network(GAN) and Autoencoder with GCN; STGNN\cite{wang2020traffic} adopts transformer (attention) for  better global/long-term temporal modeling; STG2Seq\cite{bai2019stg2seq} utilized GCN for temporal modeling, which is an interesting attempt.

\noindent\textbf{Improvement along Spatial Axis.} A lot of effort has been put on the spatial axis, that is the graph. (1) From single-graph to multi-graph. STGCN and DCRNN only use a single graph, directed or non-directed, to describe the spatial relationship. However, multimodal correlations and compound spatial dependencies exist among regions. Therefore, a series of researches elevate single-graph to multi-graph.
For instance, \cite{chai2018bike} and ST-MGCN\cite{geng2019spatiotemporal} consider spatial proximity, functional similarity, and road connectivity as mutli-graph, and so as T-MGCN\cite{lv2020temporal}; H-STGCN\cite{dai2020hybrid} takes travel time correlation matrix and shortest-path distance matrix as compound matrix; MRA-BGCN\cite{chen2020multi} builds the node-wise graph according to the road network distance, and the edge-wise graph according to the connectivity and competition. (2) From static graph to adaptive graph. Graph WaveNet\cite{wu2019graph}, TGC-LSTM\cite{cui2019traffic}, and AGCRN\cite{bai2020adaptive} adopt adaptive/learnable graph rather than a static one used in STGCN and DCRNN; DGCNN\cite{diao2019dynamic} proposes dynamic Laplacian matrix learning through tensor decomposition; SLCNN\cite{zhang2020spatio} designs Structure Learning Convolution (SLC) to dynamically learn the global/local graph structure. 
In addition to the above, attention-augmented GCN also demonstrated better performance in terms of spatial modeling in GaAN\cite{zhang2018gaan}, ASTGCN, and GMAN\cite{zheng2020gman}.

\section{Evaluation}
\subsection{Setting}
Towards the benchmark tasks listed in Section 2, we pick up some representative models and conduct comprehensive evaluations about their actual performances. Besides the deep models, we also implement two naive baselines as follows: (1) HistoricalAverage(HA). We average the corresponding values from historical days as the prediction result; (2) CopyLastStep(s). We directly copy the last one or multiple steps as the prediction result. Our experiments were performed on a GPU server with four GeForce GTX 2080Ti graphics cards.
As a benchmark evaluation, the following settings are kept the same for each model. The observation step is set to 6 for grid-based models, while the observation and prediction step are both set to 12 for graph-based models. The data ratio for training, validation, and testing is set as 7:1:2. Adam was set as the default optimizer, where the learning rate was set to 0.001 and the batch size was set to 64 by default. Mean Absolute Error is uniformly used as the loss function. The training algorithm would either be early-stopped if the validation error was converged within 10 epochs or be stopped after 200 epochs, and the best model on validation data would be saved. Root Mean Square Error (RMSE), Mean Absolute Error (MAE), and Mean Absolute Percentage Error (MAPE) are used as metrics, where zero values will be ignored.

\subsection{Effectiveness Evaluation}
The evaluation of grid-based models for single-step prediction is shown in Table \ref{tab:performance_grid}; the evaluation of graph-based models for multi-step prediction is shown in Table \ref{tab:trafficvolumn}.

\noindent\textbf{Evaluation for Grid-Based Model:} Table \ref{tab:performance_grid} shows that the state-of-the-art models did have advantages over the baselines (HA $\sim$ ConvLSTM). In particular, STDN showed better performances in general, PCRN and DeepSTN+ achieved the lowest MAE on BikeNYC-I and TaxiNYC respectively. None of these grid-based models could be acknowledged as a dominant one at the current stage. Through the experiment, we find that their main limitations are as follows: (1) ST-ResNet converts the video-like data to high-dimensional image-like data and uses a simple fusion-mechanism to handle different types of temporal dependency; (2) through the experiment, it was found PCRN took more epochs to converge and tended to cause overfitting; (3) DMVST-Net and STDN use local CNN to take grid (pixel) as computation unit, resulting in long training time; (4) DeepSTN+ utilized a fully-connected layer in $ConvPlus$ block, which would result in a big number of parameters on TaxiBJ; (4) Multitask Learning model\cite{zhang2019flow,jiang2019deepurbanevent} needs multiple data sources as the inputs, which hinders the applicability.


\noindent\textbf{Evaluation for Graph-Based Model:} Table \ref{tab:trafficvolumn} compares the prediction performances of our selected models at 15 minutes, 30 minutes, 60 minutes ahead on METR-LA, PeMS-BAY, PEMSD7M datasets. Through Table \ref{tab:trafficvolumn}, we can find that: (1) Despite the effect of time-series model LSTNet in short-term prediction, its performance would deteriorate as the horizon gets longer; (2) Almost all of the graph-based models achieved better performance than traditional methods and time-series models on all metrics, which proved that the addition of spatial information would bring substantial performance improvements; (3) Although the models' performances depended more or less on the dataset, the scores of DCRNN, Graph WaveNet, and MTGNN on all datasets ranked in the top 3, which also proved their robustness and versatility in traffic prediction tasks; (4) GMAN was found more prone to overfitting, due to which its performances on all three datasets were not as good as LSTNet; This is probably because GMAN adopted a global attention mechanism to capture the spatial dependency between each pair of nodes; (5) MTGNN and GraphWaveNet got most of the highest scores on different datasets and metrics. The self-adaptive/learnable graph demonstrated its great effectiveness for traffic prediction.

\begin{figure}[h]
	\centering
	\includegraphics[width=0.48\textwidth]{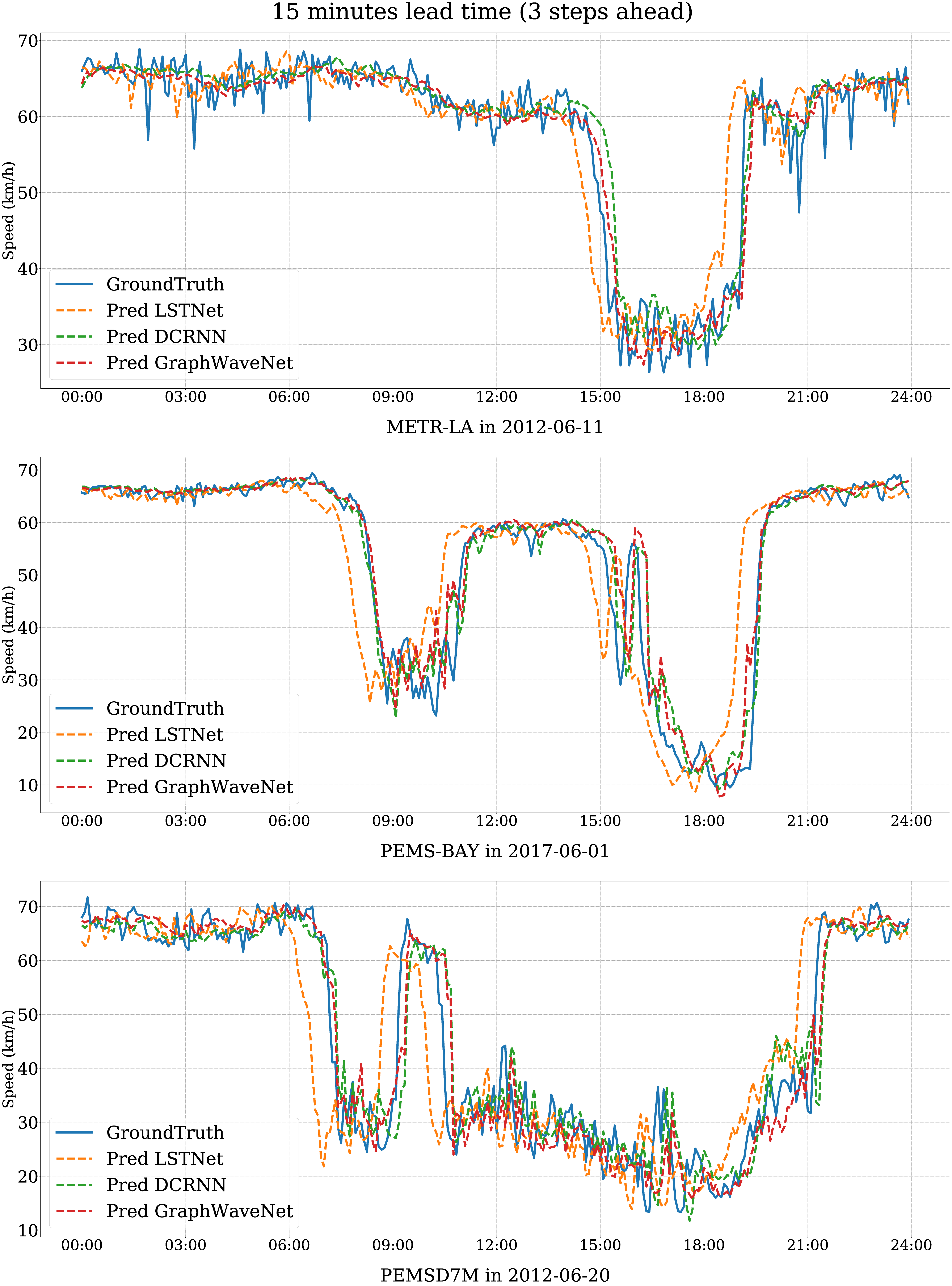}
	\caption{Case Study on Graph-Based Datasets.}
	\label{fig:Evaluation_graph}
\end{figure}

\noindent\textbf{Case Study:} We randomly selected one day (24 hours) and one sensor (node) from the three datasets (i.e., METR-LA, PEMS-BAY, and PEMSD7M) and plotted the time series of the ground truth and the predictions as Fig.\ref{fig:Evaluation_graph}. To make the time-series chart clear, in addition to the ground truth, we only plot the prediction results of LSTNet, DCRNN, and Graph WaveNet instead of all of the models listed in Table \ref{tab:trafficvolumn}. Through Fig.\ref{fig:Evaluation_graph}, we can observe that: (1) All of the three models could learn the peak and valley trend on all three datasets; (2) The graph-based models DCRNN and Graph WaveNet always outperformed the time-series model LSTNet, which proved the excellent performance of GCN in capturing spatial correlation and dependency; (3) Time lag could be observed on all of the three prediction results, especially when violent fluctuations occur in the original time series such as 2012/6/20 21:00 in PEMSD7M. This problem will be magnified in the longer-term forecast like 60 minutes lead time. Despite this, the graph-based models still show better performance in terms of volatility prediction errors, which further confirmed the effectiveness and robustness of GCN in traffic prediction.

\begin{figure}[h]
	\centering
	\includegraphics[width=0.49\textwidth]{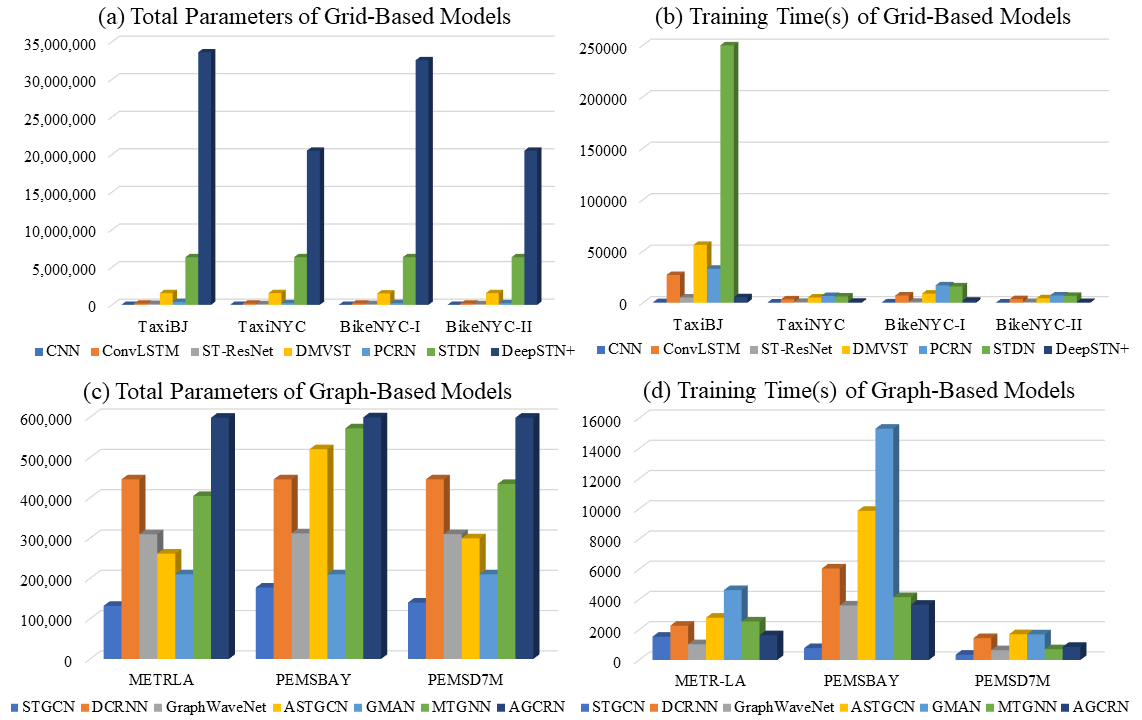}
	\caption{Efficiency Summary.}
	\label{fig:Efficiency}
\end{figure}

\subsection{Efficiency Evaluation}
Besides comparing the effectiveness of these deep models, we also provide an analysis of their efficiency. The space complexity and time complexity of these approaches are significant for their future application, so we exhibit the number of the trainable parameters and training time of each model through bar charts Fig.\ref{fig:Efficiency}. 

\noindent\textbf{Evaluation for Grid-Based Model:} From Fig.\ref{fig:Efficiency}-(a) and Fig.\ref{fig:Efficiency}-(b), we can observe that: (1) the parameter numbers of DeepSTN+ and STDN are more than other models, especially DeepSTN+, which captures the citywide spatial correlation by utilizing fully connected layer; (2) ST-ResNet has the fewest trainable parameters, and demonstrates its superiority to other models in terms of space complexity; (3) The training time of STDN and DMVST is longer than other models because they utilize the LSTM to capture the temporal dependency and take each mesh-grid as the computation unit rather than the entire mesh. 

\noindent\textbf{Evaluation for Graph-Based Model:} From Fig.\ref{fig:Efficiency}-(c) and Fig.\ref{fig:Efficiency}-(d), we can conclude that: (1) STGCN and GMAN have relatively lower space complexity than others; (2) AGCRN and DCRNN need more parameters than other models because they are based on RNNs; (3) On PEMSBAY, the parameter numbers of ASTGCN and MTGNN dramatically increase. The reason for this is those two models have more GNN layers and they are more sensitive to the node number; (4) The training time of GMAN on PEMSBAY outdistances other models because it applies a global attention mechanism to the entire graph (nodes). In summary, TCN-based models like STGCN and Graph WaveNet have higher computation efficiency.

\section{Availability and Usability}
\begin{figure}[h]
	\centering
	\includegraphics[width=0.49\textwidth]{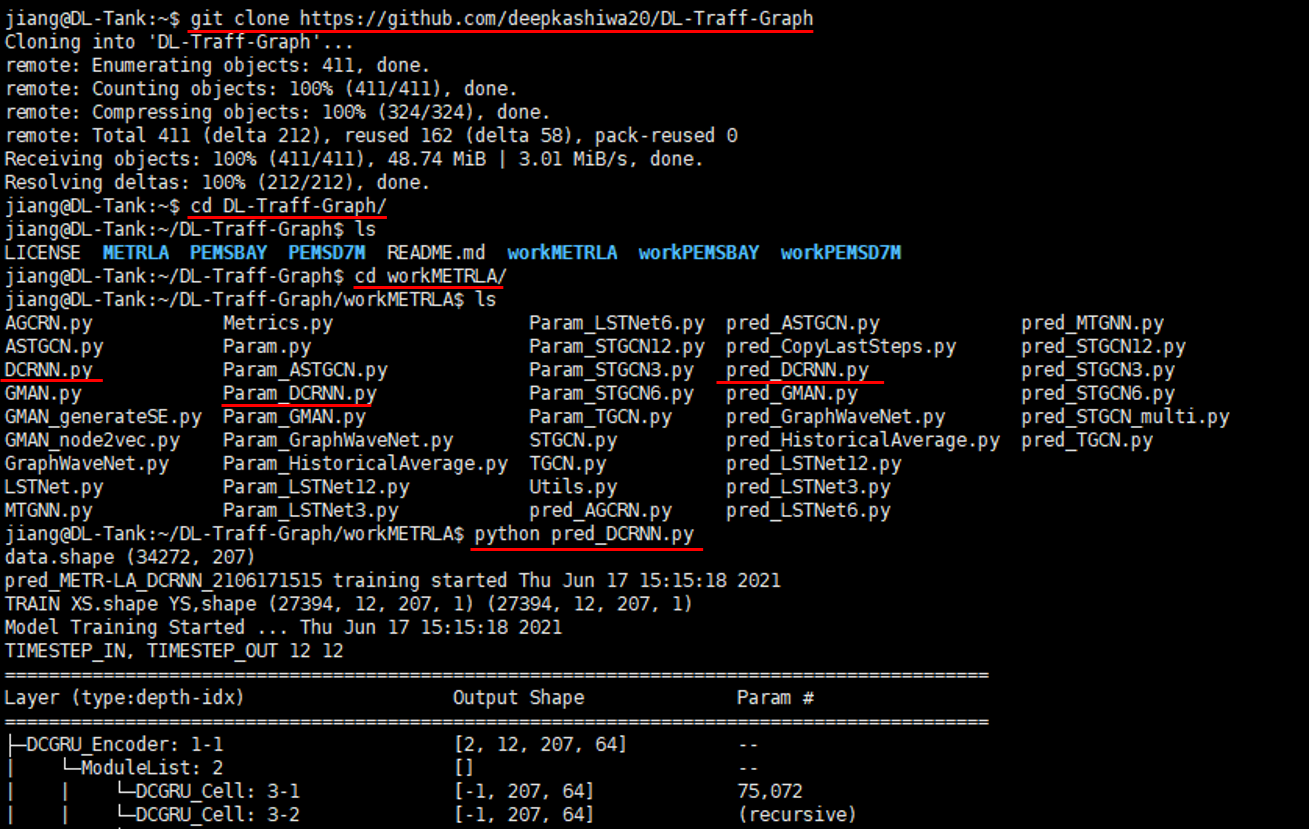}
	\caption{Illustration of The Use Case for DL-Traff.}
	\label{fig:usecase}
\end{figure}

DL-Traff is already available at GitHub as the following two repositories under the MIT License: one is for grid-based datasets/models \url{https://github.com/deepkashiwa20/DL-Traff-Grid}, and another is for graph-based datasets/models \url{https://github.com/deepkashiwa20/DL-Traff-Graph}. It is implemented with Python and the most popular deep learning frameworks: Keras\cite{keras} on TensorFlow\cite{tensorflow} and PyTorch\cite{pytorch}. Fig.\ref{fig:usecase} shows a use case by taking DCRNN model on METR-LA dataset as an example. To run the benchmark, the repository should be cloned locally and a conda environment with the necessary dependencies should be created. The directory is structured in a flat style and only with two levels. The traffic datasets are stored in DATA directories (e.g., METRLA), and the python files are put in workDATA directories (e.g., workMETRLA). Entering the work directory for a certain dataset, we can find MODEL class file (e.g., DCRNN.py) and its corresponding running program named pred\_MODEL.py (e.g., pred\_DCRNN.py). We can run ``python MODEL.py'' to simply check the model architecture without feeding the training data and run ``python pred\_MODEL.py'' to train and test the model. Additionally, Param.py file contains a variety of hyper-parameters as described in Section 5.1 that allow the experiment to be customized in a unified way. Metrics.py file contains the metric functions listed in Section 5.1. Utils.py file integrates a set of supporting functions such as pickle file reader and self-defined loss function. More details about the usability and implementation can be found at GitHub.

\section{Conclusion}
In this paper, we first survey the deep learning models as well as the widely used datasets for urban traffic prediction. Then we build a standard benchmark to comprehensively evaluate the deep traffic models on the selected open datasets. The survey and the benchmark combine together to form our study called DL-Traff, which is already available at \url{https://github.com/deepkashiwa20/DL-Traff-Grid} and \url{https://github.com/deepkashiwa20/DL-Traff-Graph}. With DL-Traff, we hope to deliver a useful and timely resource to researchers in AI and data science community.


\bibliographystyle{ACM-Reference-Format}
\bibliography{cikm2021-resource}
\end{document}